\documentclass[10pt,twocolumn,letterpaper]{article}

\usepackage{cvpr}      % To produce the REVIEW version
\usepackage{graphicx}
\usepackage{amsmath}
\usepackage{amssymb}
\usepackage{booktabs}

\usepackage{algorithm}
\usepackage{algpseudocode}
\usepackage[pagebackref,breaklinks,colorlinks]{hyperref}

\usepackage[capitalize]{cleveref}
\crefname{section}{Sec.}{Secs.}
\Crefname{section}{Section}{Sections}
\Crefname{table}{Table}{Tables}
\crefname{table}{Tab.}{Tabs.}

\def\cvprPaperID{5464} % *** Enter the CVPR Paper ID here
\def\confName{CVPR}
\def\confYear{2022}

\begin{document}

\title{Tool as Embodiment for Recursive Manipulation}

\author{Yuki Noguchi \quad Tatsuya Matsushima \quad Yutaka Matsuo\\
The University of Tokyo\\
%Matsuo Lab\\

%\and
%\\
%The University of Tokyo\\
%Matsuo Lab\\
%{\tt\small matsushima@weblab.t.u-tokyo.ac.jp}

\tt\small {\{noguchi, matsushima, matsuo\}}@weblab.t.u-tokyo.ac.jp \\

%\and
%Yutaka Matsuo\\
%The University of Tokyo\\
%Matsuo Lab\\
%{\tt\small matsuo@weblab.t.u-tokyo.ac.jp}

\and
Shixiang Shane Gu\\
Google Research\\
{\tt\small shanegu@google.com}
}
\maketitle

%%%%%%%%% ABSTRACT
\begin{abstract}
   Humans and many animals exhibit a robust capability to manipulate diverse objects, often directly with their bodies and sometimes indirectly with tools. Such flexibility is likely enabled by the fundamental consistency in underlying physics of object manipulation such as contacts and force closures. Inspired by viewing tools as extensions of our bodies, we present Tool-As-Embodiment (TAE), a parameterization for tool-based manipulation policies that treat hand-object and tool-object interactions in the same representation space. The result is a single policy that can be applied recursively on robots to use end effectors to manipulate objects, and use objects as tools, \ie new end-effectors, to manipulate other objects. By sharing experiences across different embodiments for grasping or pushing, our policy exhibits higher performance than if separate policies were trained. Our framework could utilize all experiences from different resolutions of tool-enabled embodiments to a single generic policy for each manipulation skill. Videos at \url{https://sites.google.com/view/recursivemanipulation}
\end{abstract}

% memo
% no hyphen -> end effector, gripper use, tool use, on the fly
% hypen -> closed-loop, pick-and-place, bi-manual

% unify deployment vs round -> round to stay generic
% task name
% end effector in {gripper, tool}
% tilde cite/ref
% robot vs robotic

%%%%%%%%% BODY TEXT
\section{Introduction}
\label{sec:intro}

% human animal talk?

For humans and many animals, interaction with the environment often involves the use of hands or feet. More intelligent species can even grasp and use tools to expand their range of the control over the environment~\cite{wimpenny2009cognitive,visalberghi2017cognitive,deak2014development}. Similarly, a robot system which can control both grippers and tools can empower it to handle different kinds of problems~\cite{zeng2018robotic}, especially if it is on the fly.

In existing learning paradigms for robotics, gripper use (\eg grasping~\cite{kalashnikov2018qt}, door opening~\cite{gu2017deep}, and throwing~\cite{zeng2020tossingbot}) and tool use (\eg hammering~\cite{fang2020learning} and sweeping~\cite{xie2019improvisation}) are often studied in isolation, such as works which train the tool grasp policy separately from the tool use policy~\cite{fang2020learning}.
%However, most existing robotics research only focuses on one or the other.
However, using a robot gripper or using a tool to push an object is intrinsically similar, as they share consistent laws of physics, such as friction, inertia, contacts, and force closures. In comparative psychology, it is also widely recognized that humans and animals tend to view tools as mere extensions of their own body/embodiment~\cite{cardinali2009tool}.

\begin{figure}[t]
  \centering
   \includegraphics[width=1.0\linewidth]{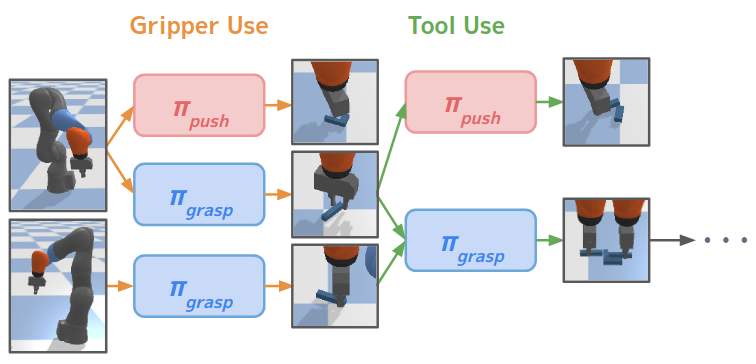}

   \caption{A recursive architecture for manipulation. Blue and red blocks respectively represent shared Grasp and Push policies. In our method, these blocks can be applied recursively to enable both Gripper Use and Tool Use. }
   \vspace{-20pt}
   \label{fig:onecol}
\end{figure}

Inspired by this phenomenon, we propose Tool-As-Embodiment (TAE) as a new learning paradigm for mastering both gripper use and tool use with a single policy.
%We adopt object-centric representations for action recently introduced~\cite{zeng2020transporter}, and make it generic regardless of if the embodiment is body or tool.
% transporters/fcns are anti-object centric -> sec 2 of transporter paper
We adopt an existing framework~\cite{zeng2020transporter} for efficient vision-based manipulation learning and make it work regardless of if the embodiment is body or tool.
The result is a universal manipulation policy that can be applied recursively. For example, using grippers to pick up objects and using those objects as tools to pick up other objects as shown in~\cref{fig:onecol}. 
% This is because...

We apply TAE to simulated KUKA iiwa robots on a series of grasping and pushing problems involving one or two arms and objects of different shapes and sizes. When presented with tasks which require both gripper use and tool use, our policy can successfully utilize all past experiences to learn more efficiently and outperforms baselines in which the policies are trained separately. We also curate and publish Recursive Manipulation (ReMa) datasets with benchmarking scores of our architecture and baseline methods, allowing future researchers to test better architectures in this unique manipulation problem. 

Our work builds toward recent works for learning a generalizable continuous control policy that is morphology-agnostic~\cite{wang2018nervenet,huang2020one
%,amorphous
} or object-agnostic~\cite{chen2021system,huang2021geometry}. Contrary to these works, our method enables diverse morphologies or embodiments through tool use, arguably a more natural way to build up rich manipulation capabilities in the real world. We hope that this can help lead to a single policy that masters all hierarchies and embodiments of manipulation.

\section{Related Work}
\label{sec:related}
% non-learning methods

\paragraph{Learning methods for a single end effector.} Deep learning methods have been shown to be applicable to robot manipulation~\cite{gu2017deep, kroemer2021review}. Grasping is a common topic, as it is often an essential step in many robotic tasks. Learning-based methods can generalize to grasp objects with various shapes and configurations~\cite{caldera2018review, mahler2017dex, zeng2018learning, pinto2016supersizing}, though they often assume a single type of gripper (\ie a parallel gripper).
%but often assume a single type of embodiment (\eg a single arm with a parallel gripper) 

A line of research in image-based robot manipulation learning relevant to our work involves the use of fully convolutional networks (FCNs). These works take advantage of dense, per-pixel calculations of convolutions and their robustness toward translation shifts of the input. Given an image of the workspace (often in top down view), the FCN outputs a dense, pixelwise action-value map, in which each output pixel corresponds to an input pixel in the same location, which in turn corresponds to a specific location in the workspace. Action-values typically represent the probability of task success in supervised learning or Q-values in reinforcement learning, and usually the action with the highest value is selected in evaluation. This approach has been used in works involving picking~\cite{zeng2018robotic, xu2020adagrasp}, pushing~\cite{zeng2018learning}, throwing~\cite{zeng2020tossingbot}, placing~\cite{zakka2020form2fit}, various other manipulation actions~\cite{zeng2020transporter, seita_bags_2021}, navigation~\cite{wu2020spatial, wu2021spatial}, and language instruction~\cite{shridhar2021cliport}. Our work follows this line with a focus on tool use.

\vspace{-10pt}

\paragraph{Generalization across different hardware.} Strong assumptions about the robotic task setup and insufficient generalization abilities of trained models remain problems in data-driven robotics. Compared to some tasks in vision~\cite{he2016deep,he2017mask} or natural language~\cite{lewis-etal-2020-bart, radford2019language}, the realization of pretrained models that work reasonably well off-the-shelf is still a challenge for robotics.

Inspired by progress catalyzed by datasets like ImageNet~\cite{krizhevsky2012imagenet}, there have been efforts to collect large datasets for robot manipulation~\cite{gupta2018robot, dasari2019robonet, mandlekar2018roboturk, jang2021bc, ebert2021bridge}. However, most do not fully address the problem of different robot hardware, and often collect data with only one type of robot model.

Other works have focused on proposing methods that generalize across different hardware settings. For example, policies conditioned on representations of agent body morphologies~\cite{wang2018nervenet, huang2020one, chen2018hardware} have been demonstrated to handle varying body types. Another way to handle embodiment differences involve training policies which can quickly adapt to changes in hardware or environment feedback using domain randomization~\cite{andrychowicz2020learning, kumar2021rma} or meta-learning~\cite{nagabandi2018learning}.
Other methods optimize not only control but also the robot hardware itself~\cite{ha2020fit2form, Xu-RSS-21}, allowing for even more flexibility and potential competence in a given task.

Some works focus specifically on the end effector. For example, UniGrasp~\cite{shao2020unigrasp} and AdaGrasp~\cite{xu2020adagrasp} train policies which can handle various 2 and 3-fingered grippers including those not seen in training. UniGrasp learns to output contact points (which correspond to fingertip locations) for stable grasps given gripper and object point clouds. AdaGrasp, which takes the FCN action-value map approach, takes advantage of cross-correlation (or ``cross convolution''~\cite{xue2016visual}) using voxel representations of the gripper and scene; our architecture also uses cross convolution but focuses on unifying gripper and tool use. 

%Fit2Form~\cite{ha2020fit2form}
%Other works also trains a grasp success predictor for different gripper shapes but also trains a generator to optimize the gripper shape given a target object.

% Lastly, there are works which transform the observation such that the policy can 
% DA type/like techniques: xirl, rl-cyclegan

\vspace{-10pt}

% tasks with tool-use
% architectures (mainly FCN-based)

% hardware agnostic
% pathak
% mask mpc

\paragraph{Learning methods for tool use.}
%Compared to direct manipulation with end effectors, there are not as many works on tool use.
 Tool use can be a powerful ability for robot manipulation as it can be viewed as on-the-fly changing of end effector hardware. However, research in data-driven learning for tool use has been relatively limited due to the extra challenges it brings on top of gripper-based control.
 
%Many assume that perception is solved or 
Action-conditional video prediction has been used to grasp and sweep with novel tools~\cite{xie2019improvisation} in a visual MPC framework~\cite{finn2017deep}.
%This method trains a single model in a visual MPC pipeline which may struggle in long-horizon .
TOG-Net~\cite{fang2020learning} trains a task-oriented grasp policy based on DexNet~\cite{mahler2017dex} by predicting the probability of task success (such as hammering or sweeping) given a grasp pose.
%which is represented through a cropped depth image.
A separate tool use policy is trained with the policy gradient algorithm. KETO~\cite{qin2020keto} learns to generate keypoints from point clouds and uses quadratic programming for tool control. Grasping is dependent on outputs from a pretrained GraspNet~\cite{mousavian20196} model.
GIFT~\cite{turpin2021gift} also learns to generate keypoints for tool use but is also dependent on a separate DexNet-like model for grasping. Lastly, other works ~\cite{toussaint2018differentiable} have shown that a combination of differentiable physics and hierarchical mixed-integer planning can optimize complex tool use trajectories without detailed reward shaping, followed up by recent work expanding its applicability~\cite{driess2021learning}.
%In contrast to these works, our method attempts to simplify the process and allow for potential feature sharing between gripper-object and tool-object manipulation.

% show convolution for encoder
\begin{figure}[t]
  \centering
   \includegraphics[width=1.0\linewidth]{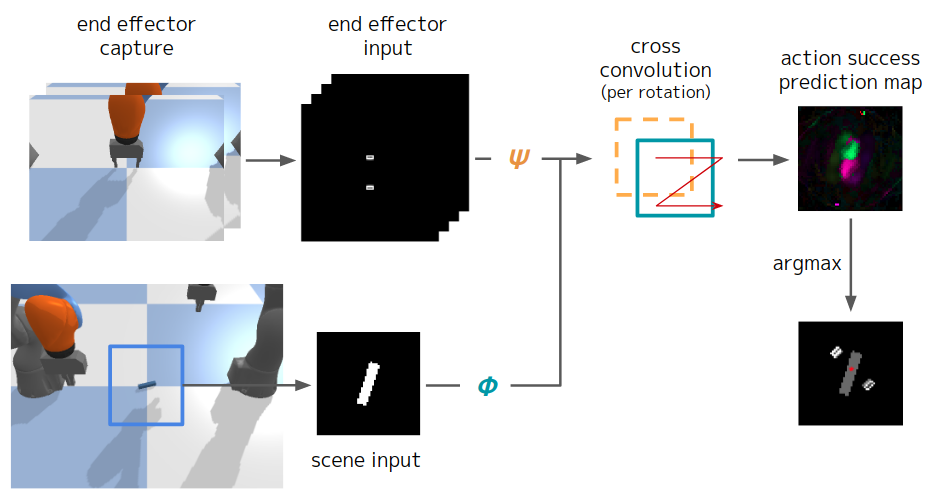}
    % variables?
   \caption{Given a manipulation task, we obtain an end effector representation and scene heightmap. Each is processed by a convolutional encoder to extract respective feature maps. The end effector features and scene features are processed together by cross convolution, and the result is decoded to produce the dense action-value map. In robot execution, the action taken corresponds to the index with the highest value in this map.}
   \vspace{-10pt}
   \label{fig:arch}
\end{figure}

\section{Tool as Embodiment}

Our method unifies gripper and tool use by representing them in the same input space~(\cref{method_inputs}). Inspired by prior work~\cite{xu2020adagrasp, zeng2020transporter}, we take advantage of fully convolutional architectures and cross-convolutions which provide an inductive bias that leads to more efficient learning~(\cref{method_arch}). The model takes as input an image representation of the end effector (``end effector'' can refer to the gripper and/or the grasped tool). and the scene (\ie the workspace) and outputs a dense action-value map in which each pixel value represents the likelihood of task success with a corresponding action. An overview of the model is visualized in~\cref{fig:arch}. We also describe how we collected the data~(\cref{method_data}) and trained the model~(\cref{method_train}) in an offline fashion, enabling easier benchmarking for future work.
%in contrast to works using similar architectures~\cite{xu2020adagrasp, zeng2020transporter} which use online learning.

% Finally, we discuss other implementation details such as data augmentation which we found to be crucial for stability (\cref{method_details}).

\begin{figure}[t]
  \centering
  \begin{subfigure}{\linewidth}
    %\fbox{\rule{0pt}{2in} \rule{.9\linewidth}{0pt}}
     \includegraphics[width=1.0\linewidth]{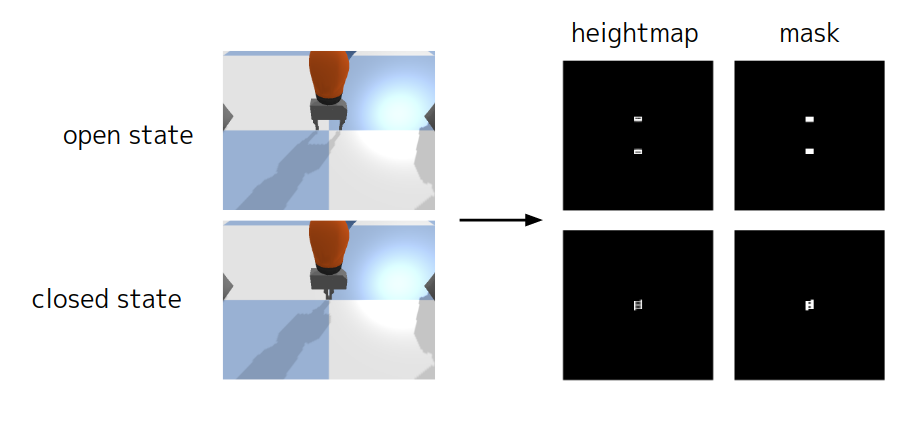}
     \caption{Parallel gripper}
    %\caption{*Success rate of learned model over multiple rounds.}
    % \label{fig:short-b}
  \end{subfigure}
  \vskip\baselineskip
  \vspace{-10pt}
  \begin{subfigure}{\linewidth}
    \includegraphics[width=1.0\linewidth]{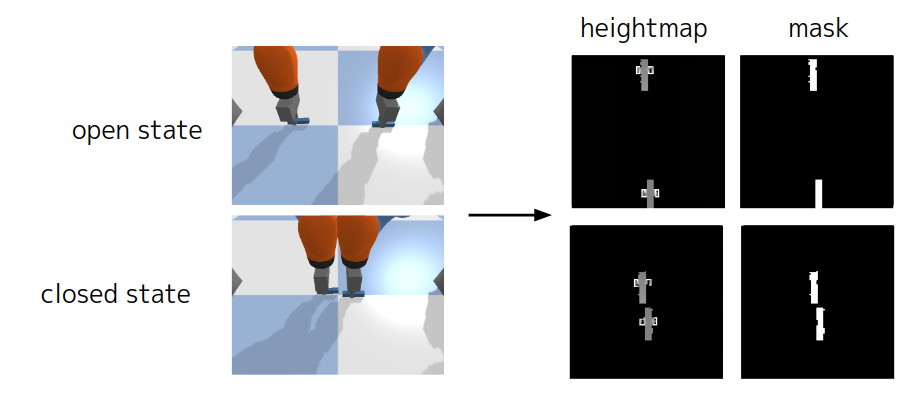}
    \caption{A pair of tools (sticks)}
  \end{subfigure}

  \caption{Examples of the proposed end effector representation (in grasping). (a) The parallel gripper (here a WSG-50) in gripper use yields simple images that looks like 2 rectangles apart and together.  (b) For tool use, an example from tool-based grasping is shown. To indicate the desire for the tools to be used and not the grippers, only they are highlighted in the masks. Note the resemblance between (a) and (b).}
  \vspace{-10pt}
  \label{fig:ee}
\end{figure}

\subsection{Unified Representation of Grippers and Tools}
\label{method_inputs}

While the scene containing the target object is represented by a single-channel heightmap $\mathbf{s} \in \mathbb{R}^{H \times W}$, end effectors are represented by a 4-channel map $\mathbf{e} \in \mathbb{R}^{H \times W \times 4}$. $\mathbf{e}$ represents 2 states of the end effector in which each state corresponds to a pair of channels. This is especially useful for grasping, in which the 2 states refer to the open and closed configuration of the gripper~\cite{xu2020adagrasp}. Examples are shown in~\cref{fig:ee}.

Each pair of channels for a state $s$ consists of a depth image $\mathbf{d}_s \in \mathbb{R}^{H \times W}$ and a mask image $\mathbf{m}_s \in \mathbb{R}^{H \times W}$, both in top down view (we use ``top down depth image'' and ``heightmap'' interchangeably). $\mathbf{d}_s$ may be obtained by capturing the end effector with one or more depth cameras, recovering a point cloud, and projecting it into a predefined plane corresponding to the workspace~\cite{zeng2020transporter}. Multiple views may be necessary if there is heavy visual occlusion of the end effector.

Mask image~$\mathbf{m}_s$ represents the areas of the end effector which is allowed to come into contact with the object to be interacted with. For example, in gripper use, pixels corresponding to the gripper fingers are assigned a value of 1 while all other pixels are 0. For tool use, we want the robot to use the tool so pixels corresponding to the tool are assigned a value of 1 while those corresponding to the gripper or anything else remain 0.
$\mathbf{m}_s$ can be obtained in a similar fashion to $\mathbf{d}_s$ using segmentation images. Segmentation images can be obtained using a variety of approaches (\eg using URDF data and forward kinematics, a learned segmentation model, etc.). In simulation, we simply access the ground truth segmentation image.

As end effector representation $\mathbf{e}$ can represent both grippers and tools, the same network, described in~\cref{method_arch}, can be used to obtain dense action-value maps for both gripper use and tool use. %the output action-value map is also shared by both.
%For example, grasping with grippers and tools require no change in 
%This means that once $\mathbf{e}$ is constructed, 

%\shane{maybe mention how action representation is also shared between gripper and tool uses? and transition to next section? ``Besides unified input space representation, we also adopt a unified output space as described in the next section"?}

\subsection{Model Architecture}
\label{method_arch}

End effector representation~$\mathbf{e}$ is processed by a convolutional encoder $\psi$ which outputs end effector features $\psi(\mathbf{e})$. Scene heightmap~$\mathbf{s}$ is similarly processed by a different convolutional encoder, $\phi$, resulting in scene features $\phi(\mathbf{s})$.

Similar to AdaGrasp~\cite{xu2020adagrasp} and Transporter~\cite{zeng2020transporter}, we use cross convolution to do feature-level matching between the scene and end effector. $\psi(\mathbf{e})$ is used as a convolutional kernel translated across $\phi(\mathbf{s})$. This results in a feature map which is further processed by a convolutional decoder that outputs the dense action-value map. Encoder and decoder architecture details are based on those of AdaGrasp, but 3D convolutions are converted to 2D convolutions for simplicity and efficiency. This is repeated $K=16$ times as the end effector image is rotated in $\theta=2\pi/K$ intervals around its center for each possible action orientation. The final stacked output is a map $Q \in \mathbb{R}^{H \times W \times K} $, where $Q(i,j,k)$ corresponds to the score of specific pose parameters for an action primitive: $ij$ corresponds to an action position and $k$ corresponds to an action orientation.

Specifically, when executing an action using the trained model, we calculate the index with the highest value in $Q$, $ijk$. $ij$ is converted back into world coordinates using the predefined workspace bounds and $k$ is multiplied by $\theta$ to recover the orientation. The recovered pose can represent the grasp pose for grasping or the start location and direction for pushing.

With a trained TAE, we can control a robot to perform gripper or tool use without switching models between different embodiments (however, note that we use separate models for different action primitives, \eg a grasp TAE and a push TAE). As a result, the policy execution process can be expressed in a recursive fashion, as visualized in~\cref{fig:onecol}.%, as shown in ~\cref{alg:tae}.

\subsection{Recursive Accumulation of Labeled Data}
\label{method_data}

As a practical way to build the dataset for our model, we collected data in discrete rounds, in which each round involves performing a set of tasks for $N$ episodes each using a random policy or a previously trained policy. A single episode of a specific task may consist of multiple steps (\eg grasp tool A $\rightarrow$ grasp tool B $\rightarrow$ grasp object C). For each step, we store observations, the sampled action, and the action outcome (mainly, a boolean indicating success). In the tasks we experiment with~(\cref{exp:tasks}), success of an action usually depends on the success of previous steps (\eg action with a tool assumes that the tool has been successfully picked up), so we terminate the episode once there is a failure at some step.

In the first round of data collection, there is no trained policy so we must start with a random policy. This random policy may have low success rate in the given tasks, but we assume that it can collect a sufficient number of successful examples for the TAE model to learn meaningful patterns.

% refine notation
This initial dataset $D_0$ is then used to train TAE model $\pi_0$. $\pi_0$ is then deployed to collect $N$ more episodes, creating ${D'}_1$. If $\pi_0$ is better than random, ${D'}_1$ would likely have more positive data than $D_0$, and this is indeed observed as described in~\cref{exp:data}. This new batch is combined with $D_0$, producing $D_1 = D_0 \cup {D'}_1$, which is used to train a new model $\pi_1$, which may have even higher performance than its predecessor $\pi_0$. This process can be repeated iteratively, creating a form of a policy improvement loop.

With this approach, we can create a dataset $D_T$, which can be shared and used as a common training dataset for comparing different methods and policies. This makes debugging and benchmarking simpler in contrast to online learning, in which it is difficult to decouple data collection and policy learning.

\subsection{Training}
\label{method_train}

The model is trained with supervised learning. Given the inputs described in~\cref{method_inputs}, the model is trained such that the index corresponding to the sampled action outputs a value of 0 or 1 depending on if the action succeeded. More details are described in prior work~\cite{xu2020adagrasp}.

During training, minibatches are sampled so that examples of success and failure are balanced on average. This is crucial because in the beginning, when there are much fewer positive examples than negative examples and the dataset is imbalanced, the model may converge to a suboptimal solution in which all values in the output are close to 0. This technique is also used in other work with similar problems~\cite{xu2020adagrasp, Mo_2021_ICCV}. We use a similar strategy for balancing data across different tasks.

%\subsection{Data augmentation}
%\label{method_details}

Additionally, we found data augmentation to be important. For example, random translation was crucial for stable outputs, possibly because convolutional layers are not completely translationally equivariant due to padding~\cite{islam2020much}.

\section{Experiments}
% describe action primitives
% multiple steps

In this section, we describe the simulation environment and tasks we created, involving both gripper use and tool use, and the dataset collected from it, which we call the ReMa (Recursive Manipulation) dataset. We also describe evaluation results for TAE on these tasks by comparing its performance to that of several baselines.

\begin{figure}
    %\fbox{\rule{0pt}{2in} \rule{.9\linewidth}{0pt}}
    \includegraphics[width=1.0\linewidth]{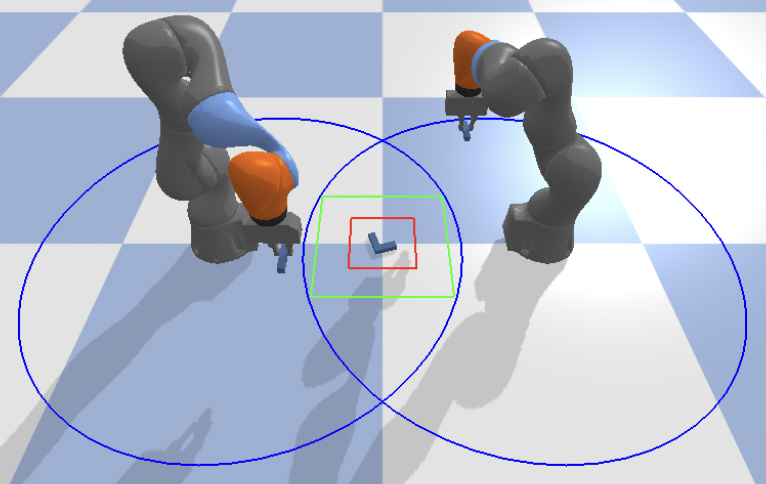}
    \caption{The environment with its 2 arms. The red square visualizes the boundaries of the scene heightmap. The green square visualizes the boundaries of the end effector representation. Each blue circle visualizes the reachable area for each arm.}
    \label{fig:field}
\end{figure}
  
\begin{figure}
    %\fbox{\rule{0pt}{2in} \rule{.9\linewidth}{0pt}}
    \includegraphics[width=1.0\linewidth]{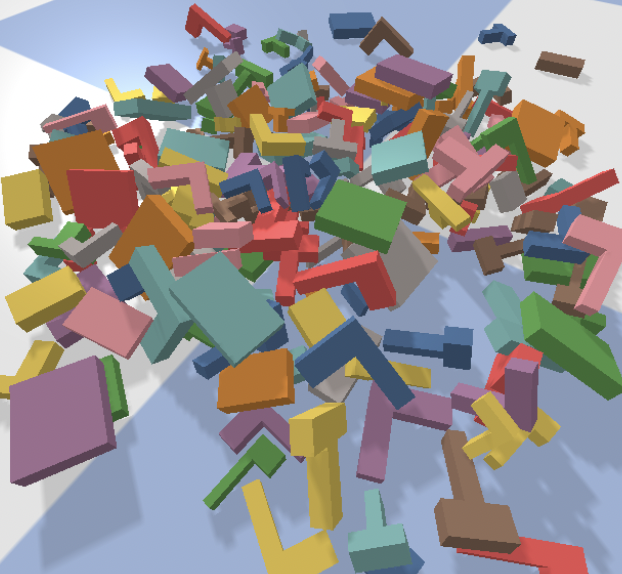}
    \caption{Samples of objects/tools used in the environment. Colors are varied for visual clarity. All samples come from one of 3 seed shapes ("L", "I", or "T") and are randomly scaled in 2 axes.}
    \label{fig:objects}
\end{figure}

\begin{figure}[t]
  \centering
  \begin{subfigure}{0.4\linewidth}
    %\fbox{\rule{0pt}{2in} \rule{.9\linewidth}{0pt}}
    \includegraphics[width=1.0\linewidth]{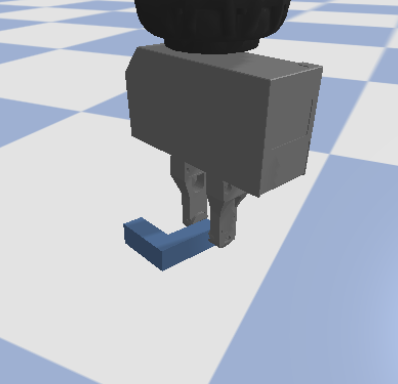}
    \caption{grasp}
    \label{fig:g-s}
  \end{subfigure}
  %\hfill
  \begin{subfigure}{0.4\linewidth}
    \includegraphics[width=1.0\linewidth]{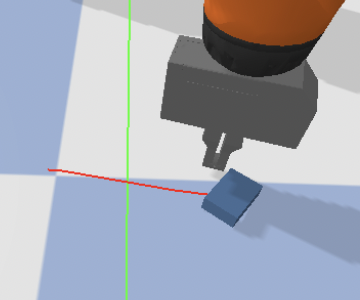}
    \caption{push}
    \label{fig:p-s}
  \end{subfigure}
  %\vskip\baselineskip
  \begin{subfigure}{0.4\linewidth}
    %\fbox{\rule{0pt}{2in} \rule{.9\linewidth}{0pt}}
     \includegraphics[width=1.0\linewidth]{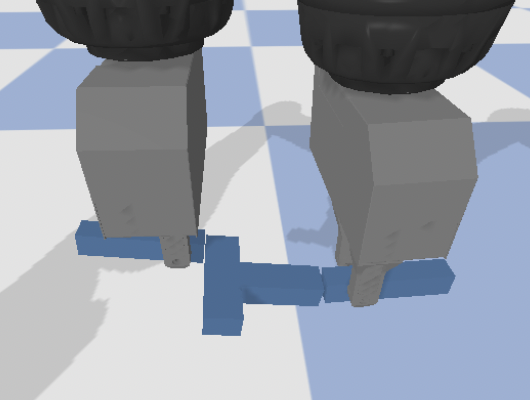}
     \caption{grasp$\rightarrow$grasp}
    \label{fig:gg-s}
  \end{subfigure}
  %\hfill
  \begin{subfigure}{0.4\linewidth}
    \includegraphics[width=1.0\linewidth]{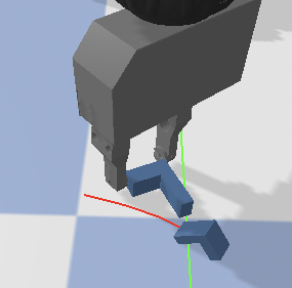}
    \caption{grasp$\rightarrow$push}
    \label{fig:gp-s}
  \end{subfigure}
  
  \begin{subfigure}{0.4\linewidth}
    \includegraphics[width=1.0\linewidth]{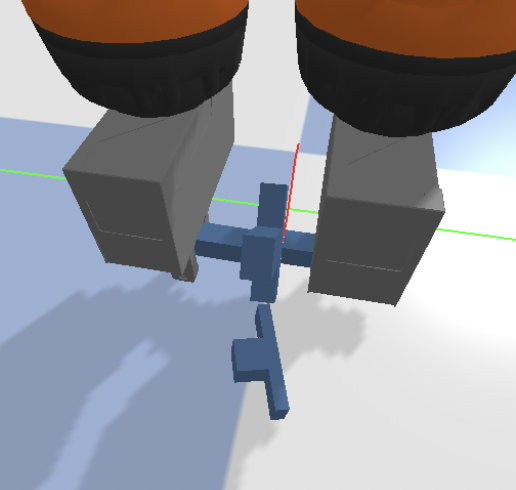}
    \caption{grasp$\rightarrow$grasp$\rightarrow$push}
    \label{fig:ggp-s}
  \end{subfigure}
  
  %\vspace{-10pt}
  \caption{Successful examples for each task. In tasks ending with pushing, the red curve visualizes the trajectory of the target object as it was pushed. The object must pass the green line for a successful push. In \emph{grasp$\rightarrow$grasp}, the target object must be lifted and held between 2 sticks~(\cref{fig:gg-s}).}
  %\vspace{-10pt}
  \label{fig:success}
\end{figure}

\subsection{Environment Setup}

%\subsubsection{Setup}

%\paragraph{Setup.}
The simulation environment consists of 2 KUKA iiwa robot arms spaced 1m apart, each with a WSG-50 gripper~(\cref{fig:field}). All target objects are in the shape of a rectangle (an "I"), an "L", or a "T"~(\cref{fig:objects}), unless otherwise stated. They are rotated and scaled randomly when spawned.

Before every action, the end effector is captured from 4 different views around the center, while the scene is captured from a single top down view. The captured images are then processed into input maps as described in~\cref{method_inputs}. Resolutions and other parameters in this process are listed in~\cref{tab:hyperparam}.

% figure of objects and setup, dimensions?

%\vspace{-10pt}

%\paragraph{Observations.}
%\vspace{-10pt}

\subsection{Tasks}
\label{exp:tasks}

Here we describe the tasks which involve 2 types of actions and 2 categories of embodiment (\ie gripper use and tool use). Examples of successful outcomes are shown in~\cref{fig:success}. In our experiments, we focus on the first 4 tasks listed (we leave \emph{grasp$\rightarrow$grasp$\rightarrow$push} out of the experiments due to difficulties in physics simulation stability).

\vspace{-10pt}

\paragraph{grasp: Grasping with a gripper} The gripper must grasp the object spawned in the workspace~(\cref{fig:g-s}).

\vspace{-10pt}

\paragraph{push: Pushing with a gripper.} The gripper, which is in a closed state, must push the object in the +x direction by at least 10cm~(\cref{fig:p-s}).

\vspace{-10pt}

\paragraph{grasp$\rightarrow$grasp: Grasping with tools.} A 3-step task. Two arms are used. Each arm sequentially grasps a stick-shaped object (the tools). If both arms successfully do so, a third object is spawned. This third object is to be grasped using the 2 tools. Only the 2 tools can touch the final object; if either gripper touches it, the task is considered to have failed~(\cref{fig:gg-s}).

\vspace{-10pt}

\paragraph{grasp$\rightarrow$push: Pushing with a tool.} A 2-step task. The gripper first grasps an object (the tool). If successful, a second object is spawned which is to be pushed in the +x direction by at least 10cm using the tool. Only the tool can touch the final object; if the gripper touches it, the task is considered to have failed~(\cref{fig:gp-s}).

\vspace{-10pt}

\paragraph{grasp$\rightarrow$grasp$\rightarrow$push: Pushing with a tool$^2$.} A 4-step task. A pushing action followed by the 3 steps in \emph{grasp$\rightarrow$grasp}~(\cref{fig:ggp-s}). \newline
% todo:update other places

%\subsection{Action primitives}

%\paragraph{Actions.}
Following prior work using FCN-based policies~\cite{zeng2020transporter}, actions in this work are executed in an open-loop fashion using action primitives. Grippers are always oriented top down, and all action types are parameterized by 3 values: an xy location and an angle about the z-axis.
%We leave more complex settings such as 6dof or closed-loop control for future work.

For grasping with a gripper at pose $p$, the gripper moves to 30cm above $p$, opens the gripper, moves down to $p$, closes the gripper, and moves back up to 30cm above $p$. Pushes with a gripper are similar to grasps, but the gripper is always closed and moves 30cm in the +x direction after reaching $p$. The pushing action in \emph{grasp$\rightarrow$push} is identical to that in \emph{push} except for the presence of a tool in the gripper.

Grasping with tools in \emph{grasp$\rightarrow$grasp} is more complex due to the use of 2 arms. The action is still represented by 3 values, but orientation is constrained by half so that the arms cannot cross, similar to another work with a bi-manual setting~\cite{ha2021flingbot}. The motions are similar to gripper grasping, except now each arm serves as a "finger" with a tool attached to each "fingertip". The arms sync and mirror each other as they move to pincer around $p$. The relative pose between the 2 grippers are fixed; learning to optimize this is left to future work. The similarity between grasping with gripper use and tool can be seen in~\cref{fig:ee}.

% Section w/ more details about task setup
% Section w/ more details about dataset, data collection
% statistics about dataset

% impl. details \eg minibatch balancing
% predict potential Qs ahead of time and answer here
% possible use-cases, related approaches to dataset

% extrapolation/generalization over shapes

% possibly more results w/ final offline dataset
% recursive architecture for control

% table evaluate w/ different shapes, configs

\subsection{Recursive Manipulation (ReMa) Dataset}
\label{exp:data}

\begin{figure}
    %\fbox{\rule{0pt}{2in} \rule{.9\linewidth}{0pt}}
    \includegraphics[width=1.0\linewidth]{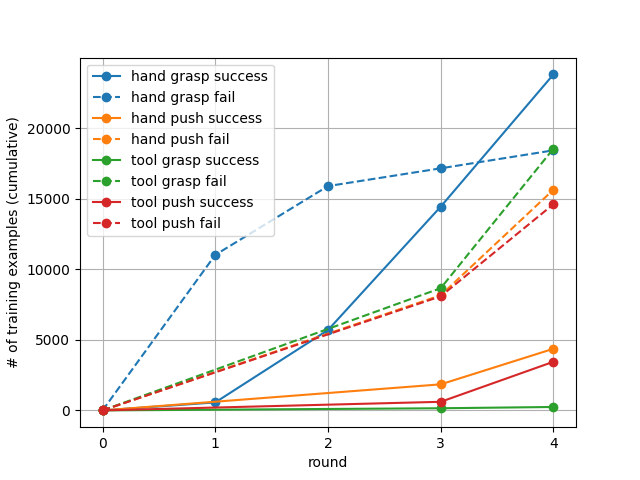}
    %*combine and align x-axis w/ other fig.
    \caption{Positive and negative data accumulated over rounds across the 4 tasks. In round 1 and 2, we only collect \emph{grasp} episodes. After grasping performance reached sufficient performance, we collected data for all 4 tasks in the next 2 rounds.}
    \label{fig:data}
  \end{figure}

%We collected $N_\tau$ episodes for each task $\tau$.
Tasks which involve tool use first require success with grasping the tool, so we trained a policy for only grasping before collecting data for other tasks. The sequence of rounds of data collection is visualized in~\cref{fig:data}.

Specifically, we first collected data for \emph{grasp} with a random policy for 12K episodes: 10K for training and 2K for validation. The resulting dataset~(\cref{fig:data}, round 1) is used to train a TAE model. This model can already achieve grasp success rates considerably higher than a random policy (6\% vs 51\%). This model is deployed to collect another 12K grasping episodes, creating a grasping dataset of size 24K~(\cref{fig:data}, round 2). At this point, there are over 5K positive and 15K negative grasp examples. A grasping model $\pi^g$ trained on this dataset produces a grasp success rate of 88\%, which we find to be sufficiently high to start collecting episodes for other tasks.

When data is collected in a tool use task for the first time, $\pi^g$ is used for grasping the tool(s). For tool use steps, a random policy is used. We produced 12K \emph{grasp$\rightarrow$grasp} episodes and 12K \emph{grasp$\rightarrow$push} episodes with this process. 12K \emph{push} episodes were also collected using a random policy. The dataset at this point corresponds to round 3 in ~\cref{fig:data}.

At this point, we have a dataset of at least 12K episodes in each of the 4 tasks. With this dataset, a new TAE model is trained from scratch. This trained model is then deployed to collect some more data. The final dataset that we obtain consists of 36K episodes for \emph{grasp} and 24K episodes each for \emph{push}, \emph{grasp$\rightarrow$grasp}, and \emph{grasp$\rightarrow$push}, totaling 108K episodes across all 4 tasks. This dataset, corresponding to round 4 in~\cref{fig:data}, is used to train TAEs in~\cref{sec:baselines}.

%Once this is trained, the model can be used to collect more data for further training and data collection to increase performance.

%\subsection{Baselines}

\begin{figure}
%\fbox{\rule{0pt}{2in} \rule{.9\linewidth}{0pt}}
    \includegraphics[width=1.0\linewidth]{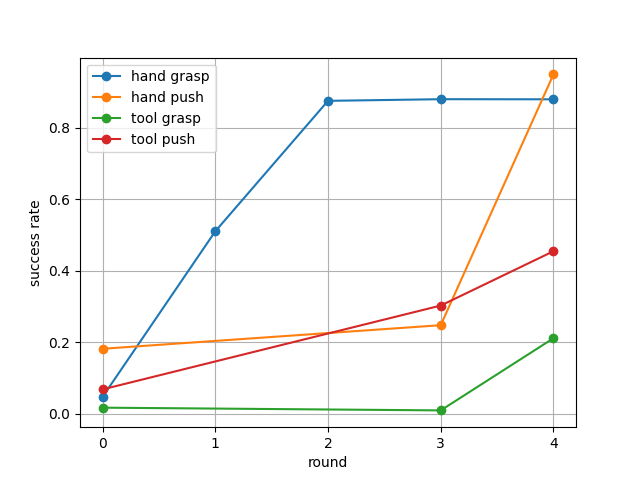}
    \caption{Success rates of jointly trained TAE over rounds across the 4 tasks. Values at round 0 correspond to success rates of a random policy. Subsequent values indicate success rates of TAE trained on the dataset in the same round in~\cref{fig:data}.}
    \label{fig:sr_curve}
% \label{fig:short-b}
\end{figure}

\begin{table}
  \centering
  \begin{tabular}{l c}
    \toprule
    hyperparameter & value \\
    \midrule
    map pixel size &  4.5mm per pixel  \\
    gripper bounds size   & 0.5m $\times$ 0.5m $\times$ 0.05m \\
    gripper map resolution  & 112 $\times$ 112 \\
    scene bounds size &  0.25m $\times$ 0.25m $\times$ 0.3m  \\
    scene map resolution  & 64 $\times$ 64  \\
    \midrule
    optimizer   & Adam \\
    learning rate &  1e-4  \\
    minibatch size &  8  \\
    \bottomrule
  \end{tabular}
  \caption{Hyperparameters used to train all TAE models.}
  \label{tab:hyperparam}
\end{table}

\begin{table}
  \centering
  \begin{tabular}{l c c c c}
    \toprule
    %method
    & grasp & push & grasp$\rightarrow$grasp & grasp$\rightarrow$push \\
    \midrule
    +ve data & 28533 & 5165 & 307 & 4118 \\
    -ve data & 22129 & 18835 & 22243 & 17540 \\
    \% +ve & 56.32 & 21.52 & 1.36 & 19.01\\
    \midrule
    random   & 5.53          & 18.23       & 1.77        & 6.87      \\
scripted & 72.77         & \emph{99.87}      & 23.47        & 22.78     \\
separate & \textbf{92.00}         & 69.05      & 18.48        & \textbf{45.96}     \\  
joint    & \textbf{93.53}         & \textbf{94.91}      & \textbf{21.17}        & \textbf{45.49}   \\
    \bottomrule
  \end{tabular}
  \caption{Recursive Manipulation (ReMa) dataset specifications and benchmarking performances. Top rows show number of examples in the final dataset accumulated from 4 rounds of data collection as described in~\cref{exp:data}. Bottom rows show success rates across the tasks for each method, each derived from the results of over 10K episodes of that task.}
  \label{tab:sr}
\end{table}

% \subsection{Learning over Multiple Embodiments}

\subsection{Comparison to Baseline Methods}
\label{sec:baselines}

We compare success rates against some baseline methods. The first baseline policy is one which samples random positions and orientations over the workspace (the red square in~\cref{fig:field}) from a uniform distribution. Orientation range is halved for bi-manual tool use grasping.

To further assess task difficulty, we also created a scripted policy with access to ground truth object positions. This policy simply outputs the target object position for grasping and adds a constant offset for pushing. Orientation is still sampled from a uniform distribution.

Results are summarized in~\cref{tab:sr}, for which we compute success rates from over 10K episodes for each method-task combination. For multi-step tool use tasks, we only evaluate episodes that made it to the last step (\ie we calculate success rate given the tools were successfully picked up). For fairer comparison against baselines in tool use tasks, we used the best learned policy for picking up tools in evaluating the baseline methods.
%so each number in Table *** is still ultimately the result of over 10K episodes.

The low performance of the random policy over all tasks indicate that they are not trivial. Even the scripted policy, which uses the ground truth position of the target object, is far from perfect (with the exception of \emph{hand push}). This indicates that position information alone is not enough to perform well in most tasks. Instead, to succeed, the geometry and interactions between the end effector and target object must be considered.

The drops in success rate of tool use tasks compared to their gripper-only counterparts for all methods indicate that the tool use tasks are more difficult. This is possibly due to the larger variety of end effector geometry that holding a tool creates and the more complex contact dynamics that emerges from tool use.

Despite the low percentage of positive data in ReMa (``\% +ve'' in~\cref{tab:sr}), TAE is able to achieve success rates higher than the data it was trained on. They are able to compete with or even surpass the scripted policy, which may indicate that they have learned meaningful features related to the end effectors and objects. Furthermore,~\cref{fig:data} and~\cref{fig:sr_curve} show a trend of increased performance with more rounds of data collection and training, suggesting success rates for TAE could go even higher.

In addition, a model that is jointly trained on both gripper and tool use is competitive with or surpasses models trained separately, suggesting that there may be some feature transfer between the embodiments, leading to better generalization. In our results, there is especially a large improvement for \emph{push}, with more than a 20 point improvement.

\begin{figure}[t]
  \centering
  \begin{subfigure}{0.4\linewidth}
    %\fbox{\rule{0pt}{2in} \rule{.9\linewidth}{0pt}}
    \includegraphics[trim={7cm 10cm 7cm 0},clip,width=1.0\linewidth]{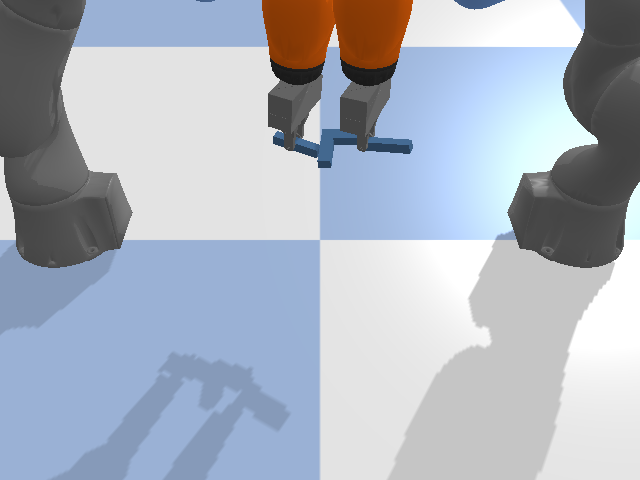}
    \caption{}
    \label{fig:fail-a}
  \end{subfigure}
  %\hfill
  \hspace{10pt}
  \begin{subfigure}{0.4\linewidth}
    \includegraphics[trim={7cm 10cm 7cm 0},clip,width=1.0\linewidth]{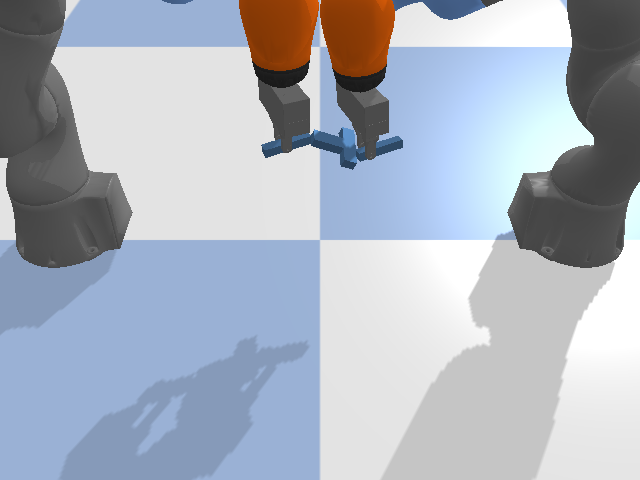}
    \caption{}
    \label{fig:fail-b}
  \end{subfigure}
  \vskip\baselineskip
  %\vspace{-10pt}
  \begin{subfigure}{0.4\linewidth}
    %\fbox{\rule{0pt}{2in} \rule{.9\linewidth}{0pt}}
    \includegraphics[trim={7cm 4cm 7cm 5cm},clip,width=1.0\linewidth]{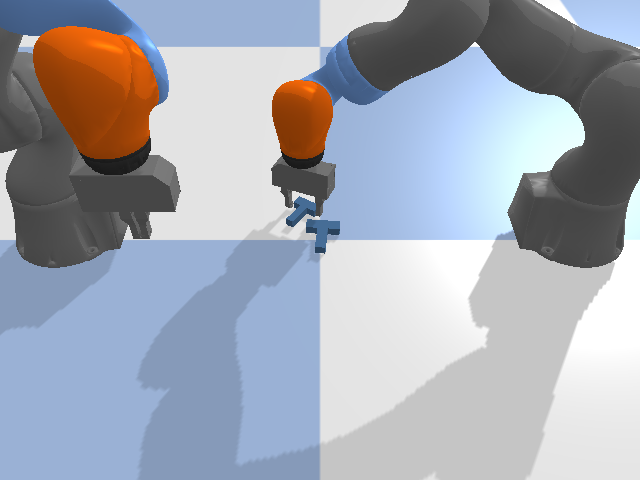}
    \caption{}
    \label{fig:fail-c}
  \end{subfigure}
  %\hfill
  \begin{subfigure}{0.4\linewidth}
    \includegraphics[trim={7cm 4cm 7cm 5cm},clip,width=1.0\linewidth]{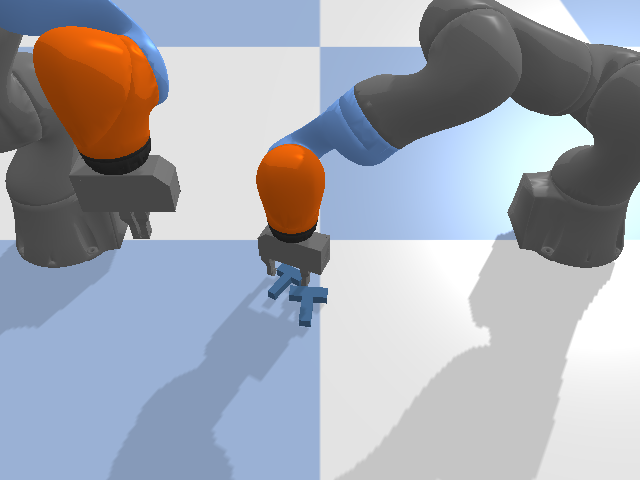}
    \caption{}
    \label{fig:fail-d}
  \end{subfigure}
  
  \caption{Examples of failure in tool use tasks. (a)(b) 2 instances of illegal contact between the gripper and the target object in \emph{grasp$\rightarrow$grasp}. (c)(d) Before and after image of a similar case in \emph{grasp$\rightarrow$push}.}
  \vspace{-20pt}
  \label{fig:fails}
\end{figure}

\subsection{Failure Cases}
\label{exp:fails}

TAE can compete with or surpass the baseline methods as described in~\cref{sec:baselines}, but its performance is still far from perfect, especially in the tool use tasks. We present one notable way it can fail in tool use tasks and discuss ways to address it.

Besides occasionally missing the target object, TAE can fail in tool use tasks from touching the object with the gripper. \cref{fig:fails} shows 3 examples of this. As described in~\cref{exp:tasks}, as a way to encourage tool use, any contact between the gripper and target object in tool use tasks is considered a failure. In real settings, this may arise in cases in which the robot is discouraged to directly handle the target due to potential harm to itself or others.

Since the mask channel in the end effector representation encodes relevant information to avoid this~(\cref{method_inputs}), collecting more training data may reduce this failure mode. However, this failure mode also relates to one important ability that we have not addressed in this work: task-oriented grasping~\cite{fang2020learning}. For example, the grasp TAE is optimized to successfully grasp the presented object/tool but does not consider how to grasp tools to maximize success in successive steps. Inspired by relevant methods~\cite{fang2020learning, zeng2018learning}, we may be able to reformulate the target value to address this (\eg replace or multiply the success label from the current action with that of the last action in the episode), but we leave this to future work.

\section{Conclusion}

We introduced Tool as Embodiment~(TAE), a robot manipulation approach for both gripper use and tool use tasks. Viewing tool use as an embodiment generalization problem, we also introduced the Recursive Manipulation benchmark, a new environment and an offline dataset featuring both grippers and tools. We have shown that TAE allows for feature sharing between embodiments, and hope that the presented work motivates more progress toward bridging the gap between body and tool.

\paragraph{Limitations and Future Work}% \shane{please fill... it can be very short}

The environment and tasks we introduce present challenging problems but also make some simplifications which should be addressed. For instance, target objects are spawned into the workspace out of thin air, one at a time before each action, which is only possible in simulation. In a more realistic setting, all objects of concern would be present from the beginning. For example, in RGB-Stacking~\cite{lee2021rgbstacking}, each episode begins with 3 color-coded objects in which one is only present as a distractor. For tool use, an interesting problem would be to present multiple potential tools and make the robot select the most appropriate one given the task.

%In addition, the objects we deal with at the moment are relatively flat shapes. 

The approach we propose builds toward generic robot policies but has limitations (besides the one discussed in~\cref{exp:fails}), some of which come from the usage of FCN-generated action-value maps. Despite its benefits in sample efficiency, it is known that methods using these often rely on 4DoF, open-loop, predefined action primitives, and our approach is no exception. Although it would come with its own set of challenges, 6DoF, closed-loop control policies would allow for more flexible and dynamic robotic behavior, which may be especially necessary for more dexterous tool use.

%The current work has following limitations; (1) only target object to manipulate is spawned into the workspace, and (2) the geometry of the objects is relatively simple. Evaluating TAE with more diverse objects and cluttered workspace would be interesting direction.
%Future work/limitations somewhere: more object diversity, multiple objects in workspace at once; \ie introduce distractors and tool selection, 6dof, closed-loop control, not relying on action primitives?, more end-to-end approach, real setting, learning hierarchical planning including tool choice

% further connect the divide between body and tool

\section*{Acknowledgement}
We thank Sergey Levine, Tom Silver, and Andy Zeng for helpful insights and discussion.

%%%%%%%%% REFERENCES
{\small
\bibliographystyle{ieee_fullname}
\bibliography{egbib}

\begin{thebibliography}{10}\itemsep=-1pt

\bibitem{andrychowicz2020learning}
OpenAI:~Marcin Andrychowicz, Bowen Baker, Maciek Chociej, Rafal Jozefowicz, Bob
  McGrew, Jakub Pachocki, Arthur Petron, Matthias Plappert, Glenn Powell, Alex
  Ray, et~al.
\newblock Learning dexterous in-hand manipulation.
\newblock {\em The International Journal of Robotics Research}, 39(1):3--20,
  2020.

\bibitem{caldera2018review}
Shehan Caldera, Alexander Rassau, and Douglas Chai.
\newblock Review of deep learning methods in robotic grasp detection.
\newblock {\em Multimodal Technologies and Interaction}, 2(3):57, 2018.

\bibitem{cardinali2009tool}
Lucilla Cardinali, Francesca Frassinetti, Claudio Brozzoli, Christian Urquizar,
  Alice~C Roy, and Alessandro Farn{\`e}.
\newblock Tool-use induces morphological updating of the body schema.
\newblock {\em Current biology}, 19(12):R478--R479, 2009.

\bibitem{chen2018hardware}
Tao Chen, Adithyavairavan Murali, and Abhinav Gupta.
\newblock Hardware conditioned policies for multi-robot transfer learning.
\newblock {\em arXiv preprint arXiv:1811.09864}, 2018.

\bibitem{chen2021system}
Tao Chen, Jie Xu, and Pulkit Agrawal.
\newblock A system for general in-hand object re-orientation.
\newblock {\em Conference on Robot Learning}, 2021.

\bibitem{dasari2019robonet}
Sudeep Dasari, Frederik Ebert, Stephen Tian, Suraj Nair, Bernadette Bucher,
  Karl Schmeckpeper, Siddharth Singh, Sergey Levine, and Chelsea Finn.
\newblock Robonet: Large-scale multi-robot learning.
\newblock {\em arXiv preprint arXiv:1910.11215}, 2019.

\bibitem{deak2014development}
Gedeon~O De{\'a}k.
\newblock Development of adaptive tool-use in early childhood: sensorimotor,
  social, and conceptual factors.
\newblock {\em Advances in child development and behavior}, 46:149--181, 2014.

\bibitem{driess2021learning}
Danny Driess, Jung-Su Ha, Russ Tedrake, and Marc Toussaint.
\newblock Learning geometric reasoning and control for long-horizon tasks from
  visual input.
\newblock In {\em Proc. of the IEEE International Conference on Robotics and
  Automation (ICRA)}, 2021.

\bibitem{ebert2021bridge}
Frederik Ebert, Yanlai Yang, Karl Schmeckpeper, Bernadette Bucher, Georgios
  Georgakis, Kostas Daniilidis, Chelsea Finn, and Sergey Levine.
\newblock Bridge data: Boosting generalization of robotic skills with
  cross-domain datasets.
\newblock {\em arXiv preprint arXiv:2109.13396}, 2021.

\bibitem{fang2020learning}
Kuan Fang, Yuke Zhu, Animesh Garg, Andrey Kurenkov, Viraj Mehta, Li Fei-Fei,
  and Silvio Savarese.
\newblock Learning task-oriented grasping for tool manipulation from simulated
  self-supervision.
\newblock {\em The International Journal of Robotics Research},
  39(2-3):202--216, 2020.

\bibitem{finn2017deep}
Chelsea Finn and Sergey Levine.
\newblock Deep visual foresight for planning robot motion.
\newblock In {\em 2017 IEEE International Conference on Robotics and Automation
  (ICRA)}, pages 2786--2793. IEEE, 2017.

\bibitem{gu2017deep}
Shixiang Gu, Ethan Holly, Timothy Lillicrap, and Sergey Levine.
\newblock Deep reinforcement learning for robotic manipulation with
  asynchronous off-policy updates.
\newblock In {\em 2017 IEEE international conference on robotics and automation
  (ICRA)}, pages 3389--3396. IEEE, 2017.

\bibitem{gupta2018robot}
Abhinav Gupta, Adithyavairavan Murali, Dhiraj Gandhi, and Lerrel Pinto.
\newblock Robot learning in homes: Improving generalization and reducing
  dataset bias.
\newblock {\em arXiv preprint arXiv:1807.07049}, 2018.

\bibitem{ha2020fit2form}
Huy Ha, Shubham Agrawal, and Shuran Song.
\newblock {Fit2Form}: 3{D} generative model for robot gripper form design.
\newblock In {\em Conference on Robotic Learning (CoRL)}, 2020.

\bibitem{ha2021flingbot}
Huy Ha and Shuran Song.
\newblock Flingbot: The unreasonable effectiveness of dynamic manipulation for
  cloth unfolding.
\newblock {\em arXiv preprint arXiv:2105.03655}, 2021.

\bibitem{he2017mask}
Kaiming He, Georgia Gkioxari, Piotr Doll{\'a}r, and Ross Girshick.
\newblock Mask r-cnn.
\newblock In {\em Proceedings of the IEEE international conference on computer
  vision}, pages 2961--2969, 2017.

\bibitem{he2016deep}
Kaiming He, Xiangyu Zhang, Shaoqing Ren, and Jian Sun.
\newblock Deep residual learning for image recognition.
\newblock In {\em Proceedings of the IEEE conference on computer vision and
  pattern recognition}, pages 770--778, 2016.

\bibitem{huang2021geometry}
Wenlong Huang, Igor Mordatch, Pieter Abbeel, and Deepak Pathak.
\newblock Generalization in dexterous manipulation via geometry-aware
  multi-task learning.
\newblock {\em arXiv preprint arXiv:2111.03062}, 2021.

\bibitem{huang2020one}
Wenlong Huang, Igor Mordatch, and Deepak Pathak.
\newblock One policy to control them all: Shared modular policies for
  agent-agnostic control.
\newblock In {\em International Conference on Machine Learning}, pages
  4455--4464. PMLR, 2020.

\bibitem{islam2020much}
Md~Amirul Islam, Sen Jia, and Neil~DB Bruce.
\newblock How much position information do convolutional neural networks
  encode?
\newblock {\em arXiv preprint arXiv:2001.08248}, 2020.

\bibitem{jang2021bc}
Eric Jang, Alex Irpan, Mohi Khansari, Daniel Kappler, Frederik Ebert, Corey
  Lynch, Sergey Levine, and Chelsea Finn.
\newblock Bc-0: Zero-shot task generalization with robotic imitation learning.
\newblock In {\em 5th Annual Conference on Robot Learning}, 2021.

\bibitem{kalashnikov2018qt}
Dmitry Kalashnikov, Alex Irpan, Peter Pastor, Julian Ibarz, Alexander Herzog,
  Eric Jang, Deirdre Quillen, Ethan Holly, Mrinal Kalakrishnan, Vincent
  Vanhoucke, et~al.
\newblock Qt-opt: Scalable deep reinforcement learning for vision-based robotic
  manipulation.
\newblock {\em arXiv preprint arXiv:1806.10293}, 2018.

\bibitem{krizhevsky2012imagenet}
Alex Krizhevsky, Ilya Sutskever, and Geoffrey~E Hinton.
\newblock Imagenet classification with deep convolutional neural networks.
\newblock {\em Advances in neural information processing systems},
  25:1097--1105, 2012.

\bibitem{kroemer2021review}
Oliver Kroemer, Scott Niekum, and George Konidaris.
\newblock A review of robot learning for manipulation: Challenges,
  representations, and algorithms.
\newblock {\em J. Mach. Learn. Res.}, 22:30--1, 2021.

\bibitem{kumar2021rma}
Ashish Kumar, Zipeng Fu, Deepak Pathak, and Jitendra Malik.
\newblock Rma: Rapid motor adaptation for legged robots.
\newblock {\em arXiv preprint arXiv:2107.04034}, 2021.

\bibitem{lee2021rgbstacking}
Alex~X. Lee, Coline Devin, Yuxiang Zhou, Thomas Lampe, Konstantinos Bousmalis,
  Jost~Tobias Springenberg, Arunkumar Byravan, Abbas Abdolmaleki, Nimrod
  Gileadi, David Khosid, Claudio Fantacci, Jose~Enrique Chen, Akhil Raju, Rae
  Jeong, Michael Neunert, Antoine Laurens, Stefano Saliceti, Federico Casarini,
  Martin Riedmiller, Raia Hadsell, and Francesco Nori.
\newblock Beyond pick-and-place: Tackling robotic stacking of diverse shapes.
\newblock In {\em Conference on Robot Learning (CoRL)}, 2021.

\bibitem{lewis-etal-2020-bart}
Mike Lewis, Yinhan Liu, Naman Goyal, Marjan Ghazvininejad, Abdelrahman Mohamed,
  Omer Levy, Veselin Stoyanov, and Luke Zettlemoyer.
\newblock {BART}: Denoising sequence-to-sequence pre-training for natural
  language generation, translation, and comprehension.
\newblock In {\em Proceedings of the 58th Annual Meeting of the Association for
  Computational Linguistics}, pages 7871--7880, Online, July 2020. Association
  for Computational Linguistics.

\bibitem{mahler2017dex}
Jeffrey Mahler, Jacky Liang, Sherdil Niyaz, Michael Laskey, Richard Doan, Xinyu
  Liu, Juan~Aparicio Ojea, and Ken Goldberg.
\newblock Dex-net 2.0: Deep learning to plan robust grasps with synthetic point
  clouds and analytic grasp metrics.
\newblock {\em arXiv preprint arXiv:1703.09312}, 2017.

\bibitem{mandlekar2018roboturk}
Ajay Mandlekar, Yuke Zhu, Animesh Garg, Jonathan Booher, Max Spero, Albert
  Tung, Julian Gao, John Emmons, Anchit Gupta, Emre Orbay, et~al.
\newblock Roboturk: A crowdsourcing platform for robotic skill learning through
  imitation.
\newblock In {\em Conference on Robot Learning}, pages 879--893. PMLR, 2018.

\bibitem{Mo_2021_ICCV}
Kaichun Mo, Leonidas~J. Guibas, Mustafa Mukadam, Abhinav Gupta, and Shubham
  Tulsiani.
\newblock Where2act: From pixels to actions for articulated 3d objects.
\newblock In {\em Proceedings of the IEEE/CVF International Conference on
  Computer Vision (ICCV)}, pages 6813--6823, October 2021.

\bibitem{mousavian20196}
Arsalan Mousavian, Clemens Eppner, and Dieter Fox.
\newblock 6-dof graspnet: Variational grasp generation for object manipulation.
\newblock In {\em Proceedings of the IEEE/CVF International Conference on
  Computer Vision}, pages 2901--2910, 2019.

\bibitem{nagabandi2018learning}
Anusha Nagabandi, Ignasi Clavera, Simin Liu, Ronald~S Fearing, Pieter Abbeel,
  Sergey Levine, and Chelsea Finn.
\newblock Learning to adapt in dynamic, real-world environments through
  meta-reinforcement learning.
\newblock {\em arXiv preprint arXiv:1803.11347}, 2018.

\bibitem{pinto2016supersizing}
Lerrel Pinto and Abhinav Gupta.
\newblock Supersizing self-supervision: Learning to grasp from 50k tries and
  700 robot hours.
\newblock In {\em 2016 IEEE international conference on robotics and automation
  (ICRA)}, pages 3406--3413. IEEE, 2016.

\bibitem{qin2020keto}
Zengyi Qin, Kuan Fang, Yuke Zhu, Li Fei-Fei, and Silvio Savarese.
\newblock Keto: Learning keypoint representations for tool manipulation.
\newblock In {\em 2020 IEEE International Conference on Robotics and Automation
  (ICRA)}, pages 7278--7285. IEEE, 2020.

\bibitem{radford2019language}
Alec Radford, Jeffrey Wu, Rewon Child, David Luan, Dario Amodei, Ilya
  Sutskever, et~al.
\newblock Language models are unsupervised multitask learners.
\newblock {\em OpenAI blog}, 1(8):9, 2019.

\bibitem{seita_bags_2021}
Daniel Seita, Pete Florence, Jonathan Tompson, Erwin Coumans, Vikas Sindhwani,
  Ken Goldberg, and Andy Zeng.
\newblock {Learning to Rearrange Deformable Cables, Fabrics, and Bags with
  Goal-Conditioned Transporter Networks}.
\newblock In {\em IEEE International Conference on Robotics and Automation
  (ICRA)}, 2021.

\bibitem{shao2020unigrasp}
Lin Shao, Fabio Ferreira, Mikael Jorda, Varun Nambiar, Jianlan Luo, Eugen
  Solowjow, Juan~Aparicio Ojea, Oussama Khatib, and Jeannette Bohg.
\newblock Unigrasp: Learning a unified model to grasp with multifingered
  robotic hands.
\newblock {\em IEEE Robotics and Automation Letters}, 5(2):2286--2293, 2020.

\bibitem{shridhar2021cliport}
Mohit Shridhar, Lucas Manuelli, and Dieter Fox.
\newblock Cliport: What and where pathways for robotic manipulation.
\newblock In {\em Proceedings of the 5th Conference on Robot Learning (CoRL)},
  2021.

\bibitem{toussaint2018differentiable}
Marc~A Toussaint, Kelsey~Rebecca Allen, Kevin~A Smith, and Joshua~B Tenenbaum.
\newblock Differentiable physics and stable modes for tool-use and manipulation
  planning.
\newblock 2018.

\bibitem{turpin2021gift}
Dylan Turpin, Liquan Wang, Stavros Tsogkas, Sven Dickinson, and Animesh Garg.
\newblock Gift: Generalizable interaction-aware functional tool affordances
  without labels.
\newblock {\em arXiv preprint arXiv:2106.14973}, 2021.

\bibitem{visalberghi2017cognitive}
Elisabetta Visalberghi, Gloria Sabbatini, Alex~H Taylor, and Gavin~R Hunt.
\newblock Cognitive insights from tool use in nonhuman animals.
\newblock 2017.

\bibitem{wang2018nervenet}
Tingwu Wang, Renjie Liao, Jimmy Ba, and Sanja Fidler.
\newblock Nervenet: Learning structured policy with graph neural networks.
\newblock In {\em International Conference on Learning Representations}, 2018.

\bibitem{wimpenny2009cognitive}
Joanna~H Wimpenny, Alex~AS Weir, Lisa Clayton, Christian Rutz, and Alex
  Kacelnik.
\newblock Cognitive processes associated with sequential tool use in new
  caledonian crows.
\newblock {\em PLoS One}, 4(8):e6471, 2009.

\bibitem{wu2020spatial}
Jimmy Wu, Xingyuan Sun, Andy Zeng, Shuran Song, Johnny Lee, Szymon
  Rusinkiewicz, and Thomas Funkhouser.
\newblock Spatial action maps for mobile manipulation.
\newblock In {\em Proceedings of Robotics: Science and Systems (RSS)}, 2020.

\bibitem{wu2021spatial}
Jimmy Wu, Xingyuan Sun, Andy Zeng, Shuran Song, Szymon Rusinkiewicz, and Thomas
  Funkhouser.
\newblock Spatial intention maps for multi-agent mobile manipulation.
\newblock In {\em IEEE International Conference on Robotics and Automation
  (ICRA)}, 2021.

\bibitem{xie2019improvisation}
Annie Xie, Frederik Ebert, Sergey Levine, and Chelsea Finn.
\newblock Improvisation through physical understanding: Using novel objects as
  tools with visual foresight.
\newblock {\em arXiv preprint arXiv:1904.05538}, 2019.

\bibitem{Xu-RSS-21}
Jie Xu, Tao Chen, Lara Zlokapa, Michael Foshey, Wojciech Matusik, Shinjiro
  Sueda, and Pulkit Agrawal.
\newblock {An End-to-End Differentiable Framework for Contact-Aware Robot
  Design}.
\newblock In {\em Proceedings of Robotics: Science and Systems}, Virtual, July
  2021.

\bibitem{xu2020adagrasp}
Zhenjia Xu, Beichun Qi, Shubham Agrawal, and Shuran Song.
\newblock Adagrasp: Learning an adaptive gripper-aware grasping policy.
\newblock {\em arXiv preprint arXiv:2011.14206}, 2020.

\bibitem{xue2016visual}
Tianfan Xue, Jiajun Wu, Katherine~L Bouman, and William~T Freeman.
\newblock Visual dynamics: Probabilistic future frame synthesis via cross
  convolutional networks.
\newblock {\em arXiv preprint arXiv:1607.02586}, 2016.

\bibitem{zakka2020form2fit}
Kevin Zakka, Andy Zeng, Johnny Lee, and Shuran Song.
\newblock Form2fit: Learning shape priors for generalizable assembly from
  disassembly.
\newblock In {\em 2020 IEEE International Conference on Robotics and Automation
  (ICRA)}, pages 9404--9410. IEEE, 2020.

\bibitem{zeng2020transporter}
Andy Zeng, Pete Florence, Jonathan Tompson, Stefan Welker, Jonathan Chien,
  Maria Attarian, Travis Armstrong, Ivan Krasin, Dan Duong, Vikas Sindhwani,
  et~al.
\newblock Transporter networks: Rearranging the visual world for robotic
  manipulation.
\newblock {\em arXiv preprint arXiv:2010.14406}, 2020.

\bibitem{zeng2020tossingbot}
Andy Zeng, Shuran Song, Johnny Lee, Alberto Rodriguez, and Thomas Funkhouser.
\newblock Tossingbot: Learning to throw arbitrary objects with residual
  physics.
\newblock {\em IEEE Transactions on Robotics}, 36(4):1307--1319, 2020.

\bibitem{zeng2018learning}
Andy Zeng, Shuran Song, Stefan Welker, Johnny Lee, Alberto Rodriguez, and
  Thomas Funkhouser.
\newblock Learning synergies between pushing and grasping with self-supervised
  deep reinforcement learning.
\newblock In {\em 2018 IEEE/RSJ International Conference on Intelligent Robots
  and Systems (IROS)}, pages 4238--4245. IEEE, 2018.

\bibitem{zeng2018robotic}
Andy Zeng, Shuran Song, Kuan-Ting Yu, Elliott Donlon, Francois~R Hogan, Maria
  Bauza, Daolin Ma, Orion Taylor, Melody Liu, Eudald Romo, et~al.
\newblock Robotic pick-and-place of novel objects in clutter with
  multi-affordance grasping and cross-domain image matching.
\newblock In {\em 2018 IEEE international conference on robotics and automation
  (ICRA)}, pages 3750--3757. IEEE, 2018.

\end{thebibliography}
}

\end{document}